# A survey of AI-generated voices and their detection

Chengzhe Sun
*Department of Computer Science and Engineering, University at Buffalo, Buffalo, New York, USA*

Tianle Yang
*Department of Linguistics, University at Buffalo, Buffalo, New York, USA, and*

Siwei Lyu
*Department of Computer Science and Engineering, University at Buffalo, Buffalo, New York, USA*

**Abstract**

The ability of artificial intelligence (AI) models to generate highly realistic human voices has advanced rapidly. These technologies power accessibility tools, virtual assistants and creative applications, but they also enable harmful uses, including impersonation, fraud and disinformation. Recent incidents of voice cloning scams targeting businesses and political leaders underscore the urgent need for robust safeguards. Unlike image and video deepfakes, the detection of synthetic voices poses unique challenges due to the complexity of phonetics, prosody and auditory perception. This survey offers a comprehensive overview of AI voice generation and detection methods, encompassing both the technical foundations and the latest state-of-the-art advances. This study also identifies key open challenges, benchmark resources and future directions to make this survey useful for future researchers.

## 1. Introduction

Using computer systems to synthesize human voices that are natural and expressive has been a long-standing goal in artificial intelligence (AI), with a wide range of applications. The history of speech synthesis spans more than five decades. Early methods relied on rule-based and concatenative synthesis, in which recorded phoneme segments were stitched together to simulate speech. In the 1990s, significant improvements in the quality of synthetic speeches were achieved with statistical parametric speech synthesis based on temporal probabilistic models, most notably hidden Markov models (HMMs), trained on real human voices (Tokuda *et al.*, 2000; Zen *et al.*, 2009). HMM-based human voice synthesis systems were widely used in commercial applications such as text-to-speech systems for navigation devices, call centers and automated customer service. However, they were only applicable in restricted settings and lacked naturalness and flexibility in handling prosody and expressive variation.

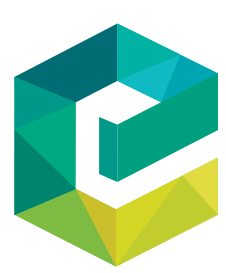

The advent of generative AI models, including generative adversarial networks (GANs), recurrent neural networks (RNNs), transformers and diffusion models, has enabled the creation of super-realistic human voices indistinguishable from real speech. The Google WaveNet model, introduced in 2016, demonstrated that neural networks could generate highly authentic waveforms, capturing subtle prosody and natural inflection (van den Oord *et al.*, 2016). Subsequent systems, such as Tacotron (Wang *et al.*, 2017) and VALL-E (Gu *et al.*, 2023), further advanced text-to-speech (TTS) and voice cloning capabilities by leveraging just a few seconds of audio. These breakthroughs have transformed synthetic voice technology into a mainstream capability, with several commercial services that now power accessibility tools, personalized virtual assistants and creative media applications.

Somehow unexpectedly, AI-generated human voices also facilitate impersonation, fraud and large-scale deception used in disinformation, undermining trust in audio evidence and communication. Numerous real-world incidents underscore the risks: in 2019, a UK energy company was defrauded when AI-based VC impersonated its parent company's CEO, resulting in a $243,000 transfer (Stupp, 2019). More recently, fraudulent robocalls that mimicked President Joe Biden's voice targeted voters during the 2024 US primary elections (Roose, 2024). The misuse of AI-generated voices highlights the pressing need to develop reliable detection methods for AI-generated human voices.

Despite multiple detection methods developed for this purpose, progress in AI-generated voice detection still lags behind the advances in detecting synthesized visual content (images and videos). This is because the audio signals possess fundamentally different characteristics from visual signals, limiting the direct applicability of image/video detection techniques to audio. To succeed, the detection of AI-generated voices must be based on a deep understanding of the unique nature of human voice signals and the mechanisms by which they are generated by AI. Consequently, understanding both how AI voices are produced and how they can be detected is a critical scientific and societal challenge.

While numerous surveys (Nguyen-Le *et al.*, 2024; Yu *et al.*, 2024; Pham *et al.*, 2025) have addressed speech synthesis and deepfake detection in isolation, this work distinguishes itself by providing an integrated, end-to-end perspective grounded in the physiology of human speech. Unlike prior reviews that focus predominantly on model architectures or low-level signal artifacts, we explicitly connect the mechanisms of AI generation (Section 3) with phonetic and articulatory constraints (Section 2) to inform robust detection strategies (Section 4). We posit that as neural vocoders increasingly minimize spectral anomalies, a deep understanding of the biological and linguistic nuances of voice production – specifically phonetics, prosody and auditory perception – is essential for identifying the next generation of deepfakes. By bridging the gap between technical generation pipelines and forensic phonetic analysis, this survey aims to provide a comprehensive roadmap for researchers addressing the unique challenges of detecting synthetic voices.

The purpose of this survey paper is to thoroughly review the current state-of-the-art in AI-generated human voices and their detection. We begin by providing a foundational background on human voice production, auditory deepfake perception, phonetics and relevant signal processing techniques (Section 2). We then provide an overview of state-of-the-art AI technologies for generating voices (Section 3), followed by a systematic review of methods for detecting synthetic speech (Section 4), highlighting the main methodologies and available benchmarking data sets. Finally, we discuss future research directions in both generation and detection of AI-generated human voices, as well as technologies beyond detection (Section 5). The literature on the synthesis and detection of AI-generated voices is vast and rapidly evolving. Although we made a concerted effort to make this survey comprehensive, it is likely that some newer work has not been covered. We aim to continuously augment the survey as new works emerge.

## 2. Physiology

Understanding the physiology behind human voice production and perception is fundamental to distinguish between natural and AI-generated voices. This section outlines two key components: the biological mechanisms related to the generation of human speech, and the auditory processes through which listeners perceive and evaluate deepfake audio. By grounding the discussion in the physical and perceptual bases of voice, we can better understand the limitations of current synthesis systems and the challenges they face in replicating the fine-grained nuances of human vocal communication.

### *2.1 Human voice production*

In most modern speech technologies, including TTS and automatic speech recognition (ASR), the phoneme functions as a basic modeling unit. Yet the precise status of the phoneme is not straightforward, since it represents an abstract category that organizes a wide range of acoustic realizations. From the perspective of speech production, phonemes can be differentiated along systematic articulatory and acoustic dimensions. Vowels, as illustrated in Table 1, are traditionally distinguished by tongue position, tongue height, lip rounding, and tenseness, dimensions that reflect continuous movements of the vocal tract and other articulators. Consonants, in contrast, are classified by their place and manner of articulation and by whether the vocal folds vibrate during their production, as summarized in Table 2. These categories structure the inventory of contrasts in English and many other languages, and they provide a principled basis for describing how speech sounds differ.

Understanding the phonetic categories of phonemes such as vowels and consonants is beneficial for deepfake audio detection. Speech synthesis systems must capture not only global acoustic distributions but also the fine-grained articulatory patterns that define these phonemic contrasts. Deviations in phoneme characteristics often occur because models cannot fully reproduce the articulatory precision of human speech. Such deviations, though subtle, provide reliable cues for distinguishing natural from synthetic voices (Agarwal *et al.*, 2020; Zhang *et al.*, 2025; Salvi *et al.*, 2025; Dhamyal *et al.*, 2021; Shi *et al.*, 2024). A phoneme-based perspective is therefore indispensable for evaluating both the accuracy and authenticity of generated speech. Moreover, because state-of-the-art deepfake models are built on TTS, voice conversion, or hybrid frameworks (Wang *et al.*, 2025), their outputs are constrained by the statistical and lexical distributions of training data. These constraints give rise to systematic weaknesses that are most clearly revealed by features anchored at the segmental (phonemic) level.

**Table 1.** North American English vowels classified by tongue position, tongue height, lip rounding and tenseness based on previous language description (Ladefoged, 1999)

| Phoneme | Tongue position | Tongue height | Lip rounding | Tenseness |
|---|---|---|---|---|
| / i / | Front | High | Unrounded | Tense |
| / I / | Front | High | Unrounded | Lax |
| / u / | Back | High | Rounded | Tense |
| / [illegible] / | Back | High | Rounded | Lax |
| / e / | Front | Mid-high | Unrounded | Tense |
| / o / | Back | Mid-high | Rounded | Tense |
| / [illegible] / | Central | Mid | Unrounded | Lax |
| / [illegible] / | Central | Mid | Unrounded | Lax |
| / Λ / | Central | Mid-low | Unrounded | Lax |
| / ε / | Front | Mid-low | Unrounded | Lax |
| / ɔ / | Back | Mid-low | Rounded | Lax |
| / æ / | Front | Low | Unrounded | Lax |
| / [illegible] / | Back | Low | Unrounded | Tense |
| / aI / | Front → Front | Low → High | Unrounded | – |
| / a[illegible] / | Front → Back | Low → High | Glide to Rounded | – |
| / ɔI / | Back → Front | Mid-low → High | Rounded → Unrounded | – |

**Table 2.** North American English consonants classified by place, manner and voicing based on previous language description (Ladefoged, 1999)

| Phoneme | Place of articulation | Manner of articulation | Voicing |
|---|---|---|---|
| / p / | Bilabial | Plosive | Voiceless |
| / b / | Bilabial | Plosive | Voiced |
| / t / | Alveolar | Plosive | Voiceless |
| / d / | Alveolar | Plosive | Voiced |
| / k / | Velar | Plosive | Voiceless |
| / g / | Velar | Plosive | Voiced |
| / t\vint / | Post-alveolar | Affricate | Voiceless |
| / d▯ / | Post-alveolar | Affricate | Voiced |
| / m / | Bilabial | Nasal | Voiced |
| / n / | Alveolar | Nasal | Voiced |
| / ŋ / | Velar | Nasal | Voiced |
| / f / | Labiodental | Fricative | Voiceless |
| / v / | Labiodental | Fricative | Voiced |
| / θ / | Dental | Fricative | Voiceless |
| / ð / | Dental | Fricative | Voiced |
| / s / | Alveolar | Fricative | Voiceless |
| / z / | Alveolar | Fricative | Voiced |
| / \vint / | Post-alveolar | Fricative | Voiceless |
| / ▯ / | Post-alveolar | Fricative | Voiced |
| / h / | Glottal | Fricative | Voiceless |
| / r / | Alveolar | Approximant | Voiced |
| / j / | Palatal | Approximant | Voiced |
| / w / | Velar | Approximant | Voiced |
| / l / | Alveolar | Lateral approximant | Voiced |

Several sources of error illustrate how these weaknesses emerge in practice. One recurrent problem is the handling of accentual and dialectal realization. TTS systems may produce frequent, in-domain words consistently, yet switch to irregular accent patterns for rare or out-of-domain items (Taylor and Richmond, 2019; He *et al.*, 2022; Zhou *et al.*, 2024). Voice conversion models typically transfer timbre and average prosody but do not reliably capture accentual targets, and accent conversion has therefore been treated as a distinct and difficult problem (Aryal and Gutierrez-Osuna, 2014; Jin *et al.*, 2023). Perceptual studies confirm that accentual and prosodic features strongly influence how listeners evaluate synthetic voices (Bakkouche *et al.*, 2025), while dialectal contrasts such as /u/-fronting in California English or the low-back merger in Midwestern English remain salient and measurable (Labov and Boberg, 2006; Hall-Lew, 2011). Many of these observed dialectal variations are caused by different means of articulation, such as the differences in tongue height or lip rounding. A second limitation is the anatomical constraints: vocal fold length and vocal tract geometry shape individual voices, yet generative models trained to minimize average error tend to regress toward population means (Ren *et al.*, 2022; Kögel *et al.*, 2023). This over-smoothing produces speech that sounds generally human but fails to capture individual-specific characteristics. Classic and contemporary findings link fundamental frequency to vocal fold physiology, vocal-tract length and body size (Titze, 1989; Pisanski *et al.*, 2014; Zhang, 2021), all of which are accessible through segment-level (phonemic-level) analysis. A third source of weakness concerns socially and linguistically conditioned phonetic variation. Features such as creaky or breathy phonation and systematic phrase-final lengthening function as markers of language, identity and style (Rose and Tianle, 2022; Podesva, 2007; Yuasa, 2010; Gordon and Ladefoged, 2001). These context-dependent variables are rarely modeled explicitly. Instead, current systems rely on coarse latent embeddings of style or prosody, which improve expressiveness but lack segment-level speaker-specific control (Zaïdi *et al.*, 2021). In sum, these issues demonstrate that attention to real human speech production is not only theoretically motivated but also practically necessary for identifying the subtle artifacts that distinguish deepfake from genuine speech.

### *2.2 Human auditory perception*

If an AI-generated voice can be reliably distinguished by human listeners, its potential to deceive or cause harm is substantially reduced. In the early stages of voice synthesis, perceptual cues such as unnatural prosody, robotic timbre, or articulatory discontinuities often made detection by ear relatively easy. However, recent advances in deep learning have significantly narrowed the perceptual gap between natural and synthetic speech. State-of-the-art models now produce voices that are not only intelligible, but often highly natural-sounding to untrained listeners. In this section, we relate the basic human perceptual system to studies of human perception of deepfake speech.

Across recent studies, unaided human detection is shown to be unreliable. In a large online experiment with English and Mandarin materials, listeners correctly spotted deepfakes about 73% of the time, and showing examples of deepfakes beforehand only slightly improved performance. No difference in detectability was found between Mandarin and English (Mai *et al.*, 2023). A multiparametric study further shows that prior information about possible exposure and the quality of the synthetic audio both affect recognition outcomes (Malinka *et al.*, 2024). In a separate series of perceptual tasks that included identity matching and judgments of naturalness for clones built with a commercial system, participants judged cloned voices to belong to the same identity as the real speaker around 80% of the time, and when asked whether a voice was AI-generated or not, they were only correct about 60% of the time (Barrington *et al.*, 2025). This shows that people are often fooled by synthetic voices, especially when the synthesis system is strong.

Context also shapes judgments. With college participants, responses varied when materials had political connotations, indicating that content framing can bias real versus fake decisions (Watson *et al.*, 2021). In political speech experiments, media modality matters. When only text is available, discernment is worse than when audio or audiovisual cues are present, and deepfakes with speech synthesized by a TTS system are harder to discern than versions performed by a voice actor (Groh *et al.*, 2024). A complementary modality study reports that people tend to perceive video deepfakes as more accurate than cheap fakes, and they are more likely to share video deepfakes than cheap or audio ones. They also found that individuals with high cognitive ability are less likely to perceive deepfakes as accurate or share them across formats (Ahmed and Chua, 2023).

Targeted cue training can help, though gains are modest. Sociolinguistically informed work has defined human discernible features such as pitch, pauses, word-initial and final stop bursts, audible breath and overall audio quality, and found these cues useful for strengthening detectors and for analyzing causal links between features and spoof labels (Khanjani *et al.*, 2024). A pre- and post-design with undergraduates showed that a short training module built on these features reduced unsure responses and improved correct classification on items that were initially uncertain (Bhalli *et al.*, 2024).

A comparison study between humans and machine detectors shows overlapping weaknesses and strengths. A game-based large-scale study found that humans and detectors tend to fail on similar attack types; native listeners had an advantage over nonnatives, while information technology experience did not confer an advantage, and older participants were more susceptible than younger participants when judging a deepfake (Müller *et al.*, 2022). However, another multi-data set evaluation has a different conclusion: it reports that humans correctly classify bona fide human audio at higher rates than several benchmark models and rely on linguistic features and intuition when performing classification, while models exhibit higher false positives and miss some low-quality or robotic items that humans correctly flag (Warren *et al.*, 2024).

Overall, the literature indicates that unaided auditory perception alone is not sufficient for reliable detection under realistic conditions. Prior information, pre-conceptions and audio quality can shift outcomes, modality changes can help or hurt depending on available cues, and short cue-focused training can reduce uncertainty but does not close the gap. These results support pairing perceptual awareness with automated safeguards in applications where deception risk is high (Mai *et al.*, 2023; Malinka *et al.*, 2024; Barrington *et al.*, 2025; Warren *et al.*, 2024).

## 3. AI generation of human voices

The AI-generated human voices, also known as speech synthesis or VC, have rapidly progressed. Early systems' robotic, monotone voices have been replaced by highly natural, expressive, and emotionally rich synthetic speech that is often difficult to tell apart from real recordings. Currently, there are three main types of AI-generated human voices: TTS, voice conversion and VC. Figure 1 illustrates the distinct architectural approaches and data flow for each synthesis method, highlighting how TTS operates purely from textual input, voice conversion transforms existing speech between speakers, and VC combines text with speaker-specific embeddings to reproduce a target voice.

### *3.1 Text-to-speech*

TTS systems have evolved from rule-based methods to large-scale generative models capable of producing natural and versatile synthetic speech. Modern pipelines typically map text, with optional prosodic or stylistic conditioning, into intermediate acoustic representations such as mel spectrograms or discrete codec tokens before waveform synthesis, or generate waveforms directly. Four central design choices govern quality and adaptability:

(1) The acoustic representation, balancing the fidelity of continuous spectra with the efficiency of discrete tokens.

(2) Temporal modeling, contrasting explicit duration control with learned monotonic alignment.

(3) The decoder architecture, ranging from high-fidelity but slow autoregressive models to faster diffusion- or GAN-based approaches.

(4) The data regime, spanning single-speaker, multi-speaker and multilingual corpora.

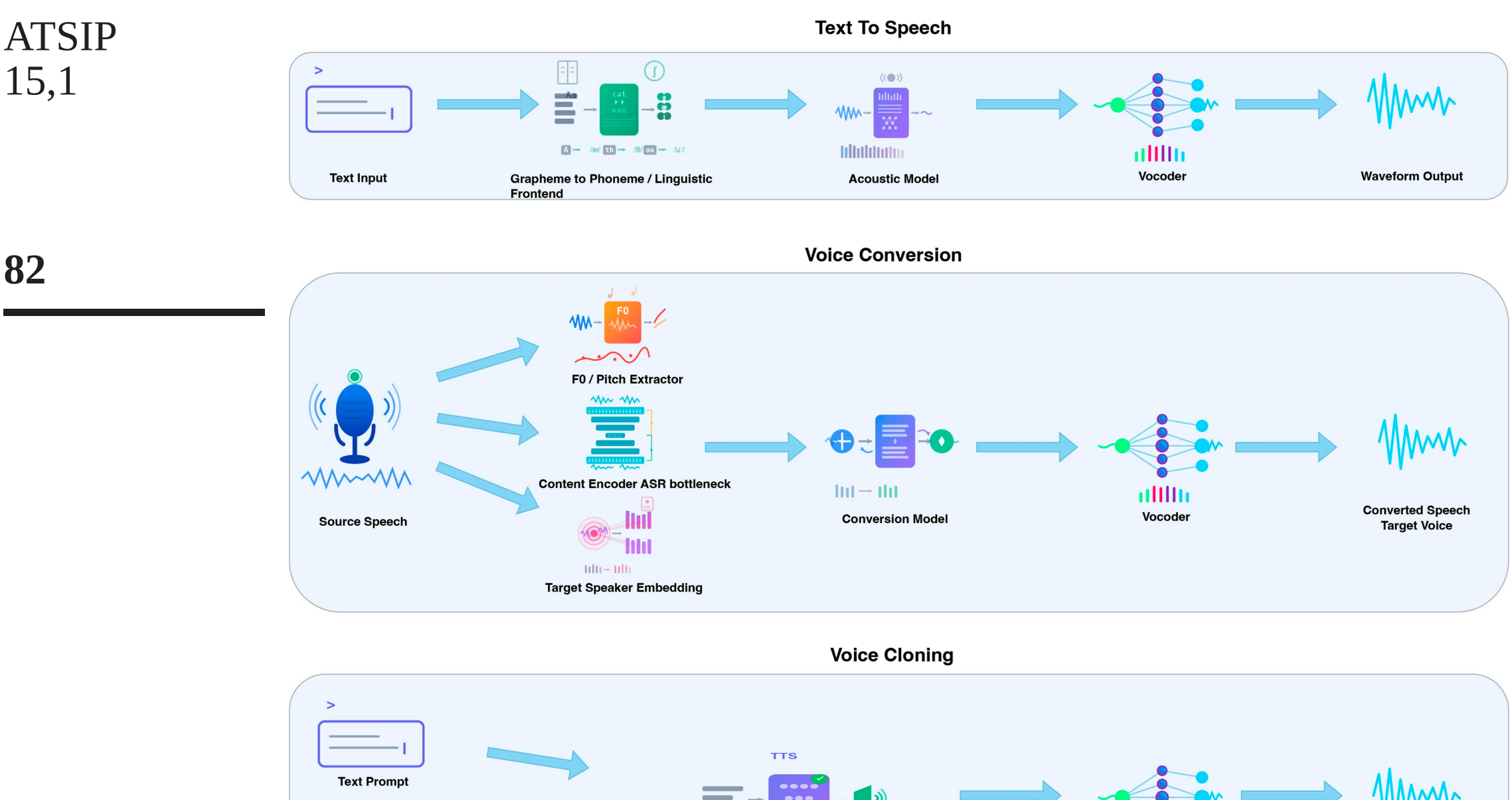

**Figure 1.** An illustration of three standard audio synthesis pipelines. Top (text-to-speech): this pipeline generates speech from text. It begins with text input, which is converted into a phonetic representation by a linguistic frontend. An acoustic model generates acoustic features from the phonemes, and a vocoder synthesizes the final waveform output. Middle (voice conversion): this pipeline transforms the voice of a source speaker into that of a target speaker. Source speech is processed to extract its linguistic content (content encoder) and pitch (F0 extractor). A conversion model combines this information with a target speaker embedding to create new acoustic features. A vocoder then synthesizes these features into the converted speech target voice. Bottom (voice cloning): this pipeline synthesizes a specific person's voice from a text prompt. A text prompt and a speaker representation from a speaker encoder are fed into a pre-trained acoustic model. A vocoder then generates the final cloned voice speech

These factors jointly determine trade-offs between naturalness, controllability, efficiency and adaptability, providing a framework for evaluating next-generation TTS models.

*3.1.1 Concatenative synthesis.* Early concatenative systems built utterances by selecting and joining recorded units such as phones, diphones or syllables to minimize linguistic and prosodic mismatch (Hunt and Black, 1996). They offered consistent intelligibility, predictable behavior and low training cost once the database was prepared. However, prosody sounded mechanical, joins were often audible, and expressiveness was limited. While techniques such as PSOLA (Moulines and Charpentier, 1990) allowed small pitch and duration modifications, larger edits introduced artifacts. Overall quality was strongly contingent on corpus coverage, personalization necessitated constructing a new database, and the upkeep of large inventories proved resource-intensive.

*3.1.2 Statistical parametric synthesis.* Statistical parametric synthesis (SPSS) used generative acoustic models, such as HMMs and Gaussian mixture models (GMMs), to predict handcrafted parameters (e.g. mel-cepstral coefficients, $F0$, and aperiodicity) (Tokuda *et al.*, 2000; Zen *et al.*, 2009). Context-dependent states enabled explicit pitch, duration and speaking-rate control, while maximum-likelihood parameter generation reduced frame jitter. Benefits included smoothness, compact models and faster adaptation than concatenative systems. The main limitation was over-smoothing, which averaged out variance, resulting in a flat timbre and restrained dynamics. Postfilters and variance compensation helped, but were unable to fully recover expressive prosody or high-frequency detail.

*3.1.3 Autoregressive attention models.* End-to-end neural TTS reframed synthesis as sequence mapping from text to acoustics with learned alignments, exemplified by Char2Wav and Tacotron, combined with neural vocoders such as WaveNet (van den Oord *et al.*, 2016). These systems captured coarticulation and natural prosody while eliminating many vocoder artifacts, achieving speaker similarity comparable to that of natural recordings. Limitations included slow autoregressive decoding, exposure bias and attention drift on long or noisy inputs, leading to skips or repetitions. Robustness improved with location-sensitive attention and guided alignments, but latency motivated the search for parallel alternatives.

*3.1.4 Parallel and integrated pipelines.* Parallel architectures addressed efficiency and stability. FastSpeech introduced a non-autoregressive decoder with a length regulator, while FastSpeech 2 added duration, pitch and energy predictors for better control (Ren *et al.*, 2019, 2021a). Glow-TTS combined monotonic alignment search with flow-based generation (Kim *et al.*, 2020). Integrated models such as VITS collapsed spectrogram prediction and vocoding into a single adversarial end-to-end system, while PortaSpeech extended this with additional control modules (Kim *et al.*, 2021; Ren *et al.*, 2021b). These approaches reduced attention failures and enabled faster inference, though sometimes at the cost of fine-grained prosody. Training remained computationally heavy for strong flow or adversarial decoders.

*3.1.5 Personalization and few-shot adaptation.* Meta-learning and style conditioning enabled rapid cloning from a few utterances. Meta StyleSpeech trained an initialization and style encoder for fast adaptation while preserving timbre and prosody (Min *et al.*, 2021). This reduced data requirements and made personalization practical, though sensitivity to recording quality and reference choice remained. Extreme accents or prosody often required additional data or fine-tuning, and cloned voices sometimes lacked the expressiveness of fully speaker-specific models.

*3.1.6 Multilingual and zero-shot generalization.* Multilingual training and speaker verification guidance allowed systems to generalize across speakers and languages without per-language engineering. YourTTS demonstrated zero-shot multilingual cloning, supporting low-resource settings and accent transfer (Casanova *et al.*, 2022). Benefits included broad coverage and flexible deployment; however, challenges included data imbalance across languages, style drift during long passages or code-switching, and sensitivity to noisy prompts.

*3.1.7 Diffusion-based TTS.* Diffusion models reformulate speech generation as an iterative denoising process conditioned on linguistic and prosodic inputs. NaturalSpeech and NaturalSpeech 2 achieved expressive and robust synthesis, even for tasks such as zero-shot singing, by combining rich conditioning with improved noise schedules (Shen *et al.*, 2022, 2023). StyleTTS 2 further integrated diffusion with adversarial training and large speech language modeling, achieving near-natural ratings (Kim *et al.*, 2023b). Building upon these advances, SupertonicTTS streamlined this paradigm by adopting a flow-matching-based latent diffusion framework, replacing iterative denoising with continuous vector field integration and eliminating external components such as G2P modules and text–speech aligners (Kim *et al.*, 2025). These diffusion-based approaches offer state-of-the-art fidelity and control over style. Despite their progress, these models still require multi-step inference and substantial computation. Few-step samplers and hybrid architectures mitigate but do not eliminate latency.

*3.1.8 Codec modeling, scale and efficiency.* Recent systems replace spectrograms with discrete codec tokens and scale training to massive corpora. BASE TTS trained billion-parameter autoregressive Transformers on over 100k hours of speech, achieving high robustness and quality but at great resource cost (Łajszczak *et al.*, 2024). VALL-E and VALL-E 2 modeled codec token sequences autoregressively, enabling near-human zero-shot cloning but with continued prompt sensitivity (Wang *et al.*, 2023; Chen *et al.*, 2024). NaturalSpeech 3 explicitly factorized content, timbre and prosody for finer control (Ju *et al.*, 2024). In parallel, models such as MaskGCT, CLaM-TTS and CM-TTS have improved inference efficiency through masked generation, constrained language models or compressed token streams, trading a slight fidelity loss for significant latency gains (Wang *et al.*, 2024; Kim *et al.*, 2024; Li *et al.*, 2024). XTTS extended zero-shot synthesis to 16 languages, highlighting demand for universal multilingual TTS but also exposing cross-language drift and prompt robustness issues (Casanova *et al.*, 2024). VITA-Audio adopts interleaved text–audio token generation with a Multiple Cross-modal Token Prediction module that emits several audio tokens per step, enabling fast and natural synthesis (Long *et al.*, 2025). DiFlow-TTS applies discrete flow matching with factorized prosody, content and acoustic tokens to achieve fast, low-latency and zero-shot speech synthesis (Nguyen *et al.*, 2025).

Concatenative systems were reliable but robotic. SPSS improved flexibility but often sounded flat. Autoregressive attention models added expressiveness at the cost of speed and stability. Parallel and integrated pipelines accelerated generation and enhanced robustness, though sometimes at the expense of fine-grained prosody. Diffusion-based systems have pushed fidelity and style control further, but they remain computationally heavy. Finally, codec- and LM-based models move the field toward universal, real-time and personalized synthesis, especially at scale. Key open challenges include reducing diffusion latency without compromising quality, enhancing robustness to noisy prompts and lengthy text, achieving faithful prosody control in zero-shot and few-shot settings, and bridging performance gaps across languages and accents. Table 3 summarizes representative models (ordered by year) and highlights the trade-offs among naturalness, efficiency and adaptability.

### *3.2 Voice conversion*

Voice conversion (VC) aims to alter a source speaker's voice so that it takes on the qualities of a target speaker while preserving the linguistic content. Most systems approach this by disentangling phonetic information from speaker-dependent cues, such as timbre, formant structure and pitch, and then resynthesizing the speech waveform using a vocoder or direct generator. VC has practical applications in personalized TTS, dubbing for film and interactive media, and assistive communication technologies:

- Representation of content and timbre through mel cepstra, phonetic posteriorgrams or more recent self-supervised embeddings.
- Training data assumptions ranging from parallel to nonparallel speech and single- to multispeaker settings.
- Strategies range from parametric vocoders and neural vocoders to direct waveform generation.

**Table 3.** Representative text-to-speech (TTS) generation models

| Name | Model/approach | Languages | Adaptation ability | Evaluation metrics | Year |
|---|---|---|---|---|---|
| Concatenative TTS (Hunt and Black, 1996) | Unit selection (diphones, syllables) | Single language | None | Intelligibility tests | 1996 |
| HMM/GMM TTS (Tokuda *et al.*, 2000; Zen *et al.*, 2009) | Statistical parametric synthesis | Single language | Limited (speaker-dependent) | MOS, MCD | 2000 |
| Char2Wav (Sotelo *et al.*, 2017) | Seq2Seq + WaveNet vocoder | English | Fine-tuning | MOS, ABX | 2017 |
| Tacotron (Wang *et al.*, 2017) | Attention-based Seq2Seq + Griffin-Lim/WaveNet vocoder | English | Fine-tuning | MOS, spectrogram analysis | 2017 |
| WaveNet (van den Oord *et al.*, 2016) | Autoregressive generative vocoder | English | None | MOS, NLL | 2016 |
| Tacotron 2 (Shen *et al.*, 2018) | Seq2Seq + WaveNet vocoder | English | Fine-tuning | MOS | 2018 |
| FastSpeech (Ren *et al.*, 2019) | Feed-forward transformer | English | Limited (fine-tuning required) | MOS, MCD | 2019 |
| Glow-TTS (Kim *et al.*, 2020) | Flow-based generative model | English | Limited (fine-tuning required) | MOS, speaker similarity | 2020 |
| FastSpeech 2 (Ren *et al.*, 2021a) | Enhanced variance modeling | English | Limited (fine-tuning required) | MOS, speaker similarity | 2020 |
| VITS (Kim *et al.*, 2021) | End-to-end variational inference + adversarial learning | English | Few-shot | MOS, speaker similarity, CER | 2021 |
| PortaSpeech (Ren *et al.*, 2021b) | Variational inference + normalizing flow | English | Few-shot | MOS, CER | 2021 |
| Meta-StyleSpeech (Min *et al.*, 2021) | Meta-learning for speaker adaptation | English | Few-shot | MOS | 2021 |
| NaturalSpeech (Shen *et al.*, 2022) | Diffusion probabilistic model | English | Few-shot | MOS, CER | 2022 |
| YourTTS (Casanova *et al.*, 2022) | Multilingual VITS-based | Multilingual | Zero-shot (cross-lingual) | MOS, speaker similarity | 2022 |
| VALL-E (Wang *et al.*, 2023) | Neural codec language model | English | Zero-shot | MOS, speaker similarity | 2023 |
| NaturalSpeech 2 (Shen *et al.*, 2023) | Latent diffusion modeling | English | Zero-shot | MOS, CER | 2023 |
| StyleTTS 2 (Kim *et al.*, 2023b) | Diffusion + adversarial training | English | Zero-shot (style transfer) | MOS, speaker similarity | 2023 |
| BASE TTS (Łajszczak *et al.*, 2024) | Large autoregressive transformer | Multilingual | Few-shot | MOS, speaker similarity, WER | 2024 |

(*continued*)

**Table 3.** Continued

| Name | Model/approach | Languages | Adaptation ability | Evaluation metrics | Year |
|---|---|---|---|---|---|
| VALL-E 2 (Chen *et al.*, 2024) | Improved neural codec LM | English | Zero-shot | MOS, speaker similarity, inference speed | 2024 |
| NaturalSpeech 3 (Ju *et al.*, 2024) | Factorized diffusion + codec modeling | Multilingual | Zero-shot | MOS, speaker similarity | 2024 |
| MaskGCT (Wang *et al.*, 2024) | Masked codec-based transformer | Multilingual | Zero-shot | MOS, inference efficiency | 2024 |
| CLaM-TTS (Kim *et al.*, 2024) | Residual vector quantization codec model | Multilingual | Zero-shot | MOS, speaker similarity | 2024 |
| CM-TTS (Li *et al.*, 2024) | Consistency model for fast inference | Multilingual | Zero-shot | MOS, inference speed (RTF) | 2024 |
| XTTS (Casanova *et al.*, 2024) | Multilingual zero-shot TTS | Multilingual | Zero-shot | MOS, speaker similarity | 2024 |
| SupertonicTTS (Kim *et al.*, 2025) | Flow-matching-based latent diffusion TTS | English | Few-shot | MOS, CER | 2025 |
| VITA-Audio (Long *et al.*, 2025) | Interleaved token generation with MCTP | English | None | MOS, inference speed (RTF) | 2025 |
| DiFlow-TTS (Nguyen *et al.*, 2025) | Discrete flow matching (factorized tokens) | English | Zero-shot | MOS, WER, RTF | 2025 |

Each of these choices involves trade-offs in terms of naturalness, controllability, efficiency and robustness, and together they define the current research challenges in building scalable, general-purpose VC systems.

*3.2.1 Statistical parametric models (GMM/HMM).* Early statistical VC using GMMs and HMMs modeled the joint distribution of aligned source-target acoustic features and applied minimum mean square error spectral mapping at conversion time (Toda *et al.*, 2007). Systems typically used STRAIGHT (Kawahara *et al.*, 1999) or WORLD (Morise *et al.*, 2016) to extract the spectral envelope, mel cepstral coefficients and aperiodicity; handled $F0$ separately through mean-variance transformation or regression; and relied on DTW or HMM-based alignments for parallel data. These methods were conceptually simple, stable with small data sets and offered interpretable components. However, they suffered from global variance shrinkage, over-smoothing, weak prosody modeling and poor robustness under channel mismatch. Despite postfilters and variance compensation, they struggled to reproduce high-frequency detail and expressive dynamics.

*3.2.2 Nonnegative matrix factorization.* Nonnegative matrix factorization (NMF) reformulated VC as spectrogram factorization $V \approx W \bullet H$, where the basis matrix $W$ captured pitch-dependent spectral templates and the activation matrix $H$ encoded their temporal usage (Nakashika *et al.*, 2013). Conversion recombined source activations with target bases. NMF improved spectral sharpness over GMM/HMM, providing an interpretable, parts-based view of timbre transfer while remaining lightweight. Limitations included only implicit modeling of temporal dynamics, reliance on Griffin-Lim or vocoders for phase reconstruction, and sensitivity in cross-speaker or cross-style scenarios, which required careful basis learning and regularization.

*3.2.3 Early neural mapping.* Neural networks replaced linear/Gaussian assumptions with multilayer perceptrons that learned nonlinear mappings from aligned parallel data (Desai *et al.*, 2009). Using stacked context windows, these models captured richer spectral relationships and reduced hand tuning compared with statistical pipelines. Still, they were data hungry, pair-specific, reliant on parallel corpora and prone to overfitting. Prosody control was weak, and alignment errors directly degraded performance.

*3.2.4 Autoencoder-based disentanglement.* Autoencoder architectures, such as AutoVC, encoded utterances into bottlenecked content codes with reduced speaker information, then conditioned a decoder on target speaker embeddings to reconstruct speech (Qian *et al.*, 2019). Content codes could be derived from acoustic encoders, ASR posteriorgrams or self-supervised features; speaker identity was obtained from learned embeddings or external d-/x-vectors. This enabled nonparallel training, zero-shot conversion to unseen speakers, and scaling to a large number of speakers with a single model. Weaknesses included flattened prosody, occasional leakage between content and timbre, sensitivity to content feature choice and dependence on robust speaker encoders. Performance degraded under noise or heavy domain shift.

*3.2.5 Adversarial and cycle-consistent models.* Adversarial and cycle-consistent frameworks removed the need for parallel data and supported flexible style transfer. CycleGAN-VC used bidirectional generators with cycle-consistency and adversarial losses to preserve content while matching timbre, and additionally incorporated identity losses to prevent over-conversion. StarGAN-based models extended this to many-to-many mapping via domain labels or style codes, with StarGAN v2-VC introducing style encoders and mapping networks for finer control (Li *et al.*, 2021). Vector quantized variants further separated content from speaker/style, aiding cross-lingual transfer. While versatile, these models were challenging to train, prone to mode collapse, and required large, diverse data sets to balance timbre transfer and strict content preservation.

*3.2.6 End-to-end waveform models.* End-to-end VC operates directly on raw audio, bypassing hand-engineered features and classical vocoders. WaveNet-based VC and newer systems, such as DRVC and NVC-Net, coupled content encoders with waveform decoders to capture fine time-frequency detail (Luo *et al.*, 2019; Wang *et al.*, 2022; Nguyen and Cardinaux, 2022). Autoregressive decoders provided high fidelity but suffered from latency, while non-autoregressive and diffusion-inspired decoders improved speed. Advantages included strong naturalness and fewer brittle preprocessing steps; disadvantages included high compute cost, larger data requirements, mismatch sensitivity and deployment challenges such as memory use, quantization and scheduling for real-time operation.

*3.2.7 Beyond identity transfer: emotion, style and multilingual VC.* Recent work targets emotion, style, accent and cross-lingual conversion. Emotional VC conditions are based on prosodic features (F0, energy, duration) or style tokens to convey emotions such as anger, joy or sadness (Zhou *et al.*, 2022). Challenges include sparse data and striking a balance between expressiveness and intelligibility. Multilingual and zero-shot VC leverage universal content spaces (e.g. posteriorgrams, self-supervised features) with speaker embeddings to enable accent and cross-lingual transfer (Jia *et al.*, 2023). These systems expand applicability but still face issues such as prosody drift, code-switching errors and degradation when training and test languages differ in phonotactics or prosodic structure.

*3.2.8 Diffusion-based VC.* Diffusion models apply iterative denoising to generate speech conditioned on content, speaker and prosody. Systems like DDDM-VC exploit strong inductive biases for denoising and integrate rich conditioning (phonetic inputs, self-supervised features, F0, rhythm) to achieve fine-grained control over timbre and expressiveness (Choi *et al.*, 2023). They deliver state-of-the-art naturalness, robust content preservation and resilience to acoustic variation. Drawbacks include slower inference due to multi-step sampling and higher training cost, though knowledge distillation and few-step samplers mitigate latency.

*3.2.9 Evaluation methodologies.* Human mean opinion score (MOS) tests remain the gold standard but are costly. Learned predictors such as MOSNet accelerate evaluation (Lo *et al.*, 2019), though they correlate imperfectly with human judgment. Consequently, evaluation triangulates subjective tests with objective metrics, including mel cepstral distortion (spectral fidelity), F0 RMSE and correlation (prosody), WER/CER via ASR (intelligibility), speaker verification error rates and cosine similarity (identity transfer) and the real-time factor (latency).

In summary, VC research has progressed from statistical models limited by over-smoothing, through adversarial and disentangled frameworks offering nonparallel and stylistic flexibility, to end-to-end and diffusion models that achieve high naturalness and control at a greater computational cost. Open challenges include reducing latency and memory use without sacrificing fidelity, improving robustness to noise and channel variability, enhancing prosody control for emotion and emphasis, and scaling reliably across speakers, styles and languages. Table 4 outlines representative systems from 2007 to 2023, summarizing their modeling strategies, data requirements, evaluation protocols and contributions. Together, these works chart the field's trajectory toward scalable, general-purpose VC.

### *3.3 Voice cloning*

Voice cloning aims to generate speech that convincingly imitates a target speaker, often using only a few reference recordings. Unlike voice conversion (VC), which transforms an existing utterance into another speaker's timbre, cloning focuses on rapid adaptation and high-fidelity similarity to the target voice. Typical pipelines consist of text analysis, an acoustic representation such as mel spectrograms or codec tokens, a speaker embedding that conditions the generator, and a vocoder or direct waveform decoder. These systems now support applications ranging from personalized assistants and dubbing to gaming, accessibility and virtual avatars.

**Table 4.** Representative voice conversion (VC) generation models

| Name | Model/approach | Languages | Evaluation metrics | Year |
|---|---|---|---|---|
| Toda *et al.* (2007) | Gaussian mixture model (GMM) | English | ABX test | 2007 |
| Desai *et al.* (2009) | Artificial neural networks (ANN) | English | MCD, MOS | 2009 |
| Nakashika *et al.* (2013) | Deep belief nets (DBNs) + concatenating NNs | English | NSD | 2013 |
| AUTOVC (Qian *et al.*, 2019) | Autoencoder bottleneck + style transfer | English | MOS | 2019 |
| Luo *et al.* (2019) | WaveNet-based VC | English | MCD | 2019 |
| AlBadawy and Lyu (2020) | VAE + GAN with WaveNet vocoder (for synthesis/detection) | English | EER, WER | 2020 |
| MaskCycleGAN-VC (Kaneko *et al.*, 2021) | Cycle-consistent adversarial training with masking | English | MCD, KDSD | 2021 |
| StarGANv2-VC (Li *et al.*, 2021) | GAN-based many-to-many framework | Multilingual | MOS, speaker similarity | 2021 |
| VQMIVC (Wang *et al.*, 2021) | Vector quantization + mutual information | English | MOS | 2021 |
| AVQVC (Tang *et al.*, 2022) | Vector quantization + AutoVC principles | English | MOS, VSS | 2022 |
| DRVC (Wang *et al.*, 2022) | Disentangled representation VC | English | MCD, MOS | 2022 |
| NVC-Net (Nguyen and Cardinaux, 2022) | End-to-end adversarial network | English | EER, MOS | 2022 |
| Jia *et al.* (2023) | Pseudo Siamese Disentanglement | Multilingual | MOS | 2022 |
| YourTTS (Casanova *et al.*, 2022) | Multilingual VITS-based VC/TTS | Multilingual | MOS, speaker similarity | 2022 |
| Emotional VC (Zhou *et al.*, 2022) | Emotional VC with ESD database | English and Chinese | MOS, ESD benchmarks | 2022 |
| DDDM-VC (Choi *et al.*, 2023) | Diffusion-based disentangled modeling | English | MOS, CER, WER, EER | 2023 |

*3.3.1 Classical approaches: unit selection and statistical synthesis.* The earliest cloning techniques were built on unit selection and SPSS. Unit selection concatenated speaker-specific segments matched to linguistic and prosodic context, while statistical methods relied on hidden Markov or GMMs to predict acoustic parameters for vocoder rendering. Once a corpus was prepared, these approaches offered stable and intelligible speech. Their limitations were significant: they required large speaker-specific data sets, produced limited expressiveness and suffered from audible artifacts such as join discontinuities (unit selection) and over-smoothing with a muffled timbre (statistical synthesis). Personalization was laborious, since each new voice demanded curating a new database.

*3.3.2 Neural TTS and fine-tuning breakthroughs.* Neural TTS models marked a turning point, enabling higher-quality cloning with far less data. Tacotron-style architectures combined with neural vocoders such as WaveNet showed that modest amounts of speaker data and fine-tuning could reproduce timbre, coarticulation and more natural prosody (Arik *et al.*, 2018; Chen *et al.*, 2019). Advantages included significant gains in naturalness and similarity, alongside a general recipe for adaptation via gradient-based updates. Drawbacks included the need for considerable computation, lengthy training, careful regularization to avoid catastrophic forgetting and per-speaker checkpoints that limited scalability.

*3.3.3 Speaker embeddings for scalable adaptation.* Scalability improved with the introduction of compact speaker embeddings. D vectors and x vectors, trained with speaker verification objectives, served as reference encoders conditioning multi-speaker acoustic or waveform models (Jia *et al.*, 2018; Wan *et al.*, 2018). These embeddings enabled the cloning of new voices from seconds of reference audio at inference time, thereby avoiding model retraining. Benefits included rapid adaptation and flexible voice swapping. Limitations included sensitivity to reference quality and channel conditions, as well as a weak capture of prosody and emotion, and degraded performance with noisy or short prompts. Moreover, embeddings risked leaking lexical or background information into the identity representation.

*3.3.4 Few-shot and meta-learning approaches.* Research then shifted toward faster, more data-efficient adaptation. Meta learning methods, such as Meta StyleSpeech, trained initializations that adapt rapidly from minimal data, while AdaSpeech explicitly modeled speaker and channel variability for robustness. These strategies enabled quicker personalization, reduced data requirements and improved stability (Min *et al.*, 2021). However, challenges remained, including training complexity, sensitivity to atypical accents or speaking styles and reduced expressiveness compared to fully fine-tuned models. In practice, careful prompt selection and occasional light fine-tuning were often still required.

*3.3.5 Zero-shot voice cloning.* Zero-shot methods removed the need for per-speaker adaptation altogether. YourTTS demonstrated that a multilingual, multi-speaker model combined with a verification encoder could synthesize unseen voices and languages from short prompts. VALL-E reframed speech as discrete codec tokens and trained a language model to generate new speech conditioned on a reference prompt (Shen *et al.*, 2022; Wang *et al.*, 2023). Advantages included strong generalization to unseen speakers, multilingual capability and the ability to clone immediately from minimal samples. Limitations included dependence on very large training corpora, uneven quality for underrepresented accents and vulnerability to noisy or mismatched prompts, which could introduce artifacts or carry background conditions into the output.

*3.3.6 Large-scale diffusion and codec models.* Recent advances combine diffusion decoders and large codec-based language models with scale. NaturalSpeech 2 and 3 integrate conditioning on content, timbre and prosody into diffusion or variational frameworks, while VALL-E 2 enhances codec modeling and long context prompting (Shen *et al.*, 2023; Chen *et al.*, 2024; Ju *et al.*, 2024). ProDiff introduced a progressive fast diffusion model that accelerates inference through adaptive step scheduling while maintaining high-quality mel-spectrogram generation, with extensions to multi-speaker scenarios (Huang *et al.*, 2022). DiffGAN-TTS combined diffusion models with adversarial training to improve sample quality and training stability for multi-speaker synthesis, enabling more efficient generation (Liu *et al.*, 2022). UnitSpeech employed discrete speech units from self-supervised models as an intermediate representation for speaker-adaptive synthesis, allowing the diffusion process to operate on more structured acoustic features for enhanced controllability in few-shot and zero-shot adaptation (Kim *et al.*, 2023a). XTTS (Cross-lingual TTS) extended zero-shot cloning to multilingual scenarios using a VQ-VAE encoder with GPT-based autoregressive modeling and diffusion-based decoding, achieving robust voice transfer across 16 languages with minimal reference audio (Casanova *et al.*, 2023). Commercial systems, such as ElevenLabs, have also emerged, deploying advanced neural architectures optimized for production use with an emphasis on real-time synthesis, VC from short samples, and cross-lingual capabilities (ElevenLabs, 2023). These systems achieve near-human similarity, finer style control, robustness for long prompts and real-time or near-real-time synthesis using efficient samplers. Their drawbacks are steep computational and data requirements, large inference footprints, brittleness under domain shift and ethical risks, given how convincingly they can clone voices from very short samples.

*3.3.7 Evaluation metrics and safeguards.* Evaluation and safeguards are increasingly integral. Human listening tests remain the gold standard, while automatic metrics provide scale. Common measures include speaker verification similarity scores and equal error rate (EER) for identity transfer, word or character error rate for intelligibility, F0 error and correlation for prosody, spectral distortion for timbre, and real-time factor for efficiency. Automatic metrics are efficient but imperfect proxies for perception, often missing rhythm, emotion or emphasis. Robust evaluations, therefore, combine subjective tests with multiple objective measures. In parallel, safeguards such as consent workflows, watermarking and provenance tracking are being explored to mitigate misuse.

Voice cloning has evolved from resource-heavy unit selection and statistical synthesis to neural fine-tuning, embedding-based models, few-shot and meta learning, and now zero-shot cloning with diffusion and codec-based generators. Classical methods offered stability but limited expressiveness. Early neural models improved naturalness but demanded per-speaker training. Embedding-based systems unlocked rapid adaptation yet struggled with noisy prompts and emotional nuance. Few-shot and meta learning reduced data costs but still required careful references. Zero-shot and large generative models now provide broad generalization and near-human similarity, but require immense resources and raise new ethical challenges. Table 5 summarizes these model families, their methods, advantages, disadvantages, evaluation practices and publication years, highlighting open problems in prosody control, robustness under low-resource and noisy conditions, and responsible deployment.

## 4. Detection of AI-generated voices

Current detection of AI-generated voices follows two main tracks: signal-level and phonetic/linguistic analysis. Signal methods inspect acoustic traces such as spectrogram statistics (e.g. mel/linear band energy ratios), phase and group-delay irregularities, harmonic-noise balance, micro-jitter/shimmer, pitch and formant stability, and vocoder "fingerprints" left by autoregressive, GAN, or diffusion vocoders. They are fast, model-agnostic, and do not require transcripts, but are brittle under post-processing (such as compression, noise and room impulse responses), domain shift, and intentional obfuscation (including band-limiting and re-recording). Phonetic methods examine what is spoken and how, including phoneme duration and coarticulation patterns, formant trajectories across phones, prosody (F0 contour, rhythm and stress), disfluencies and higher-order consistency between text and acoustics. These cues are more semantically grounded and harder to mimic, which ideally improves interpretability; however, they depend on accurate ASR/forced alignment, degrade on short or noisy clips and vary by language, dialect and speaker context. In practice, hybrid systems that fuse calibrated signal features with phonetic/prosodic evidence and are trained for robustness to codecs and replay yield the best generalization, although zero-shot clones and unseen TTS still pose challenges.

**Table 5.** Representative voice cloning and zero/few-shot TTS models

| Name | Model/approach | Cloning type | Evaluation metrics (as reported) | Year |
|---|---|---|---|---|
| Tacotron 2 + WaveNet (Shen *et al.*, 2018) | Seq2Seq acoustic model + autoregressive vocoder | Fine-tuning (speaker adaptation) | MOS, speaker similarity (Sim) | 2018 |
| d-vector TTS (Jia *et al.*, 2018) | Speaker embeddings (d-vector) + Tacotron 2 | Few-shot | MOS, speaker similarity (Sim) | 2018 |
| x-vector TTS (Wan *et al.*, 2018) | x-vector embeddings + neural vocoder | Few-shot | MOS, speaker similarity (Sim) | 2018 |
| AdaSpeech (Chen *et al.*, 2021) | Adaptive TTS with meta-learning | Few-shot | MOS, Mel Cepstral Distortion (MCD) | 2021 |
| Meta-StyleSpeech (Min *et al.*, 2021) | Meta-learning + style adaptation | Few-shot | MOS, speaker similarity (Sim) | 2021 |
| YourTTS (Casanova *et al.*, 2022) | VITS-based, multilingual + adversarial training | Zero-shot | MOS, speaker similarity (Sim) | 2022 |
| ProDiff (Huang *et al.*, 2022) | Progressive fast diffusion model | Multi-speaker TTS | MOS, real-time factor (RTF) | 2022 |
| DiffGAN-TTS (Liu *et al.*, 2022) | Diffusion + adversarial training | Multi-speaker TTS | MOS, speaker similarity (Sim) | 2022 |
| VALL-E (Wang *et al.*, 2023) | Neural codec language model | Zero-shot | MOS, speaker similarity (Sim) | 2023 |
| NaturalSpeech 2 (Shen *et al.*, 2023) | Diffusion-based TTS | Zero-shot | MOS, CER/WER, speaker similarity (Sim) | 2023 |
| UnitSpeech (Kim *et al.*, 2023a) | Discrete units + diffusion modeling | Few-shot/zero-shot | MOS, speaker similarity (Sim) | 2023 |
| XTTS (Casanova *et al.*, 2023) | VQ-VAE + GPT + diffusion decoder | Zero-shot | MOS, speaker similarity (Sim) | 2023 |
| ElevenLabs (2023) | Commercial neural TTS system | Zero-shot | MOS (reported) | 2023 |
| NaturalSpeech 3 (Ju *et al.*, 2024) | Factorized codec + diffusion modeling | Zero-shot | MOS, WER, speaker similarity (Sim) | 2024 |
| VALL-E 2 (Chen *et al.*, 2024) | Improved codec-based transformer (successor of VALL-E) | Zero-shot | MOS, speaker similarity (Sim) | 2024 |

In the following sections, we elaborate on existing methods in these directions.

### *4.1 Signal-based detections*

As synthetic speech grows increasingly indistinguishable from natural voices, reliable detection has become critical to security and trust. Early advances were driven by the ASVspoof challenges, which provided standardized data sets, protocols and metrics for logical- and physical-access scenarios, and promoted joint reporting of speaker verification and anti-spoofing performance (e.g. EER and tandem DCF) (Wu *et al.*, 2015; Todisco *et al.*, 2019). The first generation of detectors drew directly from traditional spoofing countermeasures; hand-crafted cepstral features, such as LFCCs and CQCCs, were modeled using GMMs. These systems were lightweight and interpretable, and they performed well on the specific attack families represented in the training data. However, they also showed strong sensitivity to corpus artifacts, suffered under codec or channel variation, and generalized poorly to unseen synthesis methods.

*Spectrogram-based CNN and RNN detectors*: As neural TTS and VC matured, deep learning became the default. Spectrogram-based convolutional and recurrent models learned local spectral “micro-patterns” and longer temporal dependencies that feature-engineered systems often missed (AlBadawy and Lyu, 2020; Lavrentyeva *et al.*, 2019). AASIST advanced this line with spectro-temporal attention and graph reasoning, yielding stronger robustness across data sets and attack types (Jung *et al.*, 2021). In return for markedly higher in-domain accuracy and the flexibility to absorb data augmentation or auxiliary objectives, these models demanded more data and compute, were harder to interpret, and could still falter when microphones, sampling rates, languages or vocoders differed from those seen during training.

*Raw-waveform models*: In parallel, raw-waveform detectors dispensed with fixed front-ends. Architectures such as RawNet2 and RawNet3 learn filters directly in the time domain and stack residual blocks to model long contexts (Tak *et al.*, 2021a). By avoiding hand-crafted features, they can pick up subtle waveform-level fingerprints and often generalize better to some unseen attacks. The trade-off is practical: they are sensitive to sample-rate and channel mismatch unless heavily augmented, consume more memory and require longer computation for long utterances. They can be harder to train stably without careful regularization and curriculum design.

*Vocoder-aware detection*: A complementary direction exploits the near-ubiquity of neural vocoders in modern TTS and VC. Because decoders such as WaveNet, WaveRNN, MelGAN, Parallel WaveGAN, WaveGrad or DiffWave tend to imprint faint but systematic artifacts, detectors that learn those “fingerprints” can generalize more gracefully (Sun *et al.*, 2023). Rawnet2-vocoder instantiated this idea with a RawNet2-based multitask system that jointly classified real versus fake and identified the underlying vocoder on the LibriSeVoc corpus (Sun *et al.*, 2023). By forcing the network to recognize generator-specific cues, the system achieved strong cross-attack accuracy and proved resilient to standard post-processing such as resampling or additive noise. Its limitations mirror its strengths: coverage matters (novel decoders or codec-LM models may erode gains), shallow fingerprints can be obfuscated, and the auxiliary heads add complexity that complicates calibration in the wild.

*Challenges and emerging strategies*: Despite steady progress, several challenges persist. Domain shift remains the dominant failure mode: models trained on one collection or channel can underperform on real-world audio with room reverberation, handset artifacts or aggressive compression. At the same time, diffusion- and codec-LM-based TTS narrows acoustic gaps to human speech, diminishing the salience of low-level artifacts and making purely acoustic cues less reliable, even for humans. Looking ahead, the most promising strategies combine signal-level analysis with higher-level evidence, including phonetic and lexical consistency via ASR, prosodic and rhythmic coherence across long windows, and dialogue-level behavior. Self-supervised pretraining for robust front-ends, multitask and open-set formulations that tag attack attributes (e.g. vocoder family or generator), and better score calibration across languages and channels will likely help. In parallel, complementary defenses-watermarking, provenance and signing, and consent-aware workflows should play a larger role alongside detection.

In summary, countermeasures have evolved with generation itself: from feature-based GMMs that were simple and efficient yet brittle to unseen attacks, to spectrogram-based CNN and RNN detectors that offer high accuracy but need data and careful domain coverage, to raw-waveform networks that learn expressive filters at higher computational cost, and now to vocoder-aware systems that trade added complexity for better cross-attack generalization. Table 6 consolidates representative models, their core strategies, typical strengths and weaknesses, evaluation benchmarks and years of introduction.

### *4.2 Phonetics-based detection*

Deepfake audio detection has largely relied on general acoustic or spectral cues, yet advances in speech synthesis have made these broad differences increasingly subtle. This suggests that it may be worthwhile to explore a linguistically informed perspective, particularly from phonetics. Phonetic analysis could potentially reveal fine-grained segmental and suprasegmental features of speech that remain difficult for current synthesis models to replicate consistently (Mittal *et al.*, 2024; Yang *et al.*, 2025a; Blue *et al.*, 2022; Sivaraman *et al.*, 2025; Mallinson *et al.*, 2024; Yang *et al.*, 2025b).

Although phonetics-based approaches to deepfake detection remain relatively underexplored, recent research demonstrates their potential value. For example, Yang *et al.* (2025a) show that segmental cues, particularly vowel formants (F1–F3), can distinguish real from synthetic speech more effectively than global acoustic measures such as long-term formant distributions, long-term $F0$ or speaker-level MFCCs. These findings offer a new insight: fine-grained phonetic features, closely tied to articulatory mechanisms, capture subtle deviations that current synthesis models fail to reproduce, suggesting a promising direction for more interpretable and robust detection methods. From the articulatory phonetics perspective, Blue *et al.* (2022) introduce a detection framework that reconstructs the vocal tract configuration to identify inconsistencies between natural and synthetic speech. The method achieves high accuracy, with detection rates above 90% even on short utterances, demonstrating that current deepfake systems fail to replicate the physiological constraints of human speech production. In addition, Cho *et al.* (2023) demonstrate that self-supervised speech recognition models implicitly learn articulatory representations closely aligned with human tongue and lip movements, suggesting that such physically grounded

**Table 6.** Detection of AI-generated voices

| Name | Features/inputs | Model/approach | Attack types covered | Evaluation data set(s) | Key metrics | Year |
|---|---|---|---|---|---|---|
| CQCC-GMM (Todisco *et al.*, 2019) | CQCC (handcrafted) | Gaussian mixture model | VC, statistical TTS | ASVspoof 2015/2019 | EER, t-DCF | 2019 |
| LFCC-GMM (Todisco *et al.*, 2019) | LFCC (handcrafted) | Gaussian mixture model | VC, statistical TTS | ASVspoof 2019 | EER, t-DCF | 2019 |
| LFCC-LCNN (Lavrentyeva *et al.*, 2019) | | | LFCC + spectrogram | Light CNN | VC, TTS (logical access) | |
| ASVspoof 2019/2021 | EER, t-DCF | 2019 | | | | |
| RawNet2 (Tak *et al.*, 2021a) | Raw waveform | CNN-GRU end-to-end | TTS, VC, replay | ASVspoof 2019/2021 | EER | 2021 |
| AASIST (Jung *et al.*, 2021) | | | Spectrogram + graph features | Spectro-temporal graph attention network | VC, TTS, replay (logical/physical access) | |
| ASVspoof 2021 | EER, t-DCF | 2021 | | | | |
| RawNet3 (Tak *et al.*, 2021a) | Raw waveform | Enhanced CNN-GRU | TTS, VC, replay | ASVspoof 2021 | EER | 2021 |
| WavLM (Chen *et al.*, 2022a) | Self-supervised embeddings | Transformer-based SSL model | TTS, VC, unseen neural synthesis | LibriSeVoc, WaveFake, ASVspoof 2021 | EER, accuracy | 2022 |
| XLS-R (Babu *et al.*, 2021) | Cross-lingual embeddings | Self-supervised multilingual transformer | TTS, VC (cross-lingual) | WaveFake, ASVspoof 2021 | EER, accuracy | 2021 |
| RawNet2 Vocoder (Sun *et al.*, 2023) | Raw | | waveform + vocoder ID | Multitask RawNet2 | Neural vocoder traces | |
| LibriSeVoc, WaveFake, ASVspoof 2019 | EER, robustness tests | 2023 | | | | |

features could enhance the robustness and interpretability of deepfake speech detection systems.

In terms of the interaction among the segments, a recent study by Yang *et al.* (2025b) is perhaps the first to propose that interactions among segmental features can influence the detailed realization of speech in neural synthesis. It introduces a segmental-level prosodic probing framework that evaluates whether neural TTS systems can reproduce consonant-induced F0 perturbation, a fine-grained phonetic effect linking consonantal voicing and laryngeal tension to vowel pitch. By comparing TTS models trained on identical data, the study shows that models reproduce expected perturbation patterns for high-frequency words but fail to generalize these effects to low-frequency or unseen words. This suggests that current TTS architectures rely on surface-form memorization rather than abstract articulatory-acoustic encoding. The work further demonstrates that such missing segmental-prosodic effects can serve as diagnostic cues for detecting synthetic speech, offering a linguistically grounded framework that connects phonetic analysis, interpretability and deepfake detection.

For consonant acoustics, Sivaraman *et al.* (2025) investigate the contribution of voiced and unvoiced regions of speech to audio deepfake detection. The results imply that unvoiced sounds, particularly fricatives and stops, contain distinctive artifacts that current synthesis models fail to reproduce, and that exploiting the complementarity of voiced and unvoiced cues can enhance both the accuracy and interpretability of deepfake detection systems. From a sociolinguistics point of view, Mallinson *et al.* (2024) argue that current automatic methods are brittle and overlook the importance of linguistic variation, and they highlight how phonetic, phonological and sociolinguistic insights can augment detection systems. They also emphasize the value of training human listeners to improve perceptual discernment and propose educational strategies alongside computational approaches. The implication is that integrating linguistic expertise not only strengthens technical detection methods but also broadens societal resilience against deception, while creating opportunities for interdisciplinary collaboration between speech technology and the language sciences.

In sum, these findings indicate that a phonetic perspective on deepfake detection is both promising and insufficiently studied. Because phonetic features are closely tied to articulatory mechanisms and physiological constraints, they provide a theoretically grounded basis for identifying artifacts that synthesis models struggle to reproduce. At the same time, their explicit connection to perceptible categories enhances interpretability in forensic settings, where transparent reasoning is as important as statistical accuracy. The integration of sociophonetic insights further highlights how variation in accent, dialect and voice quality can expose systematic weaknesses in current models, offering a dimension of analysis largely absent from conventional neural acoustic approaches. By drawing on articulatory, perceptual and sociolinguistic knowledge, phonetics thus offers a complementary and potentially more powerful perspective for improving the robustness and explanatory depth of audio deepfake detection.

### *4.3 Data sets and benchmarks*

Much of the progress in audio deepfake detection has come from shared data sets and community challenges. These resources not only supply the large curated material needed for training, but also establish agreed-upon protocols and metrics so that competing systems can be compared on equal footing. The earliest benchmarks popularized EER as a standard threshold-independent measure. Later editions expanded the evaluation toolkit with metrics that reflect deployment more realistically: the tandem detection cost function (tDCF), which evaluates a spoofing countermeasure together with an ASV backend, and calibration-oriented measures such as minDCF and $C_{llr}$, which capture operating point risk and the reliability of system scores. The advantage of these shared resources is clear. The results are reproducible and directly comparable. However, they also encourage "benchmark overfitting," where models exploit quirks or attack fingerprints in the data set that fail to carry over to real-world conditions.

*ASVspoof 2015.* This first large step established a unified testbed for spoofing against ASV, with roughly 260,000 total utterances (approximately 26,000 genuine and 234,000 spoofed) from 106 speakers, mixing bona fide speech with fakes generated by classical voice conversion and statistical TTS (Wu *et al.*, 2015). Strengths: a consistent protocol that catalyzed research, an accessible data scale for rapid iteration, and the widespread adoption of EER as a community metric. Limitations: coverage focused on vocoder-based, pre-neural attacks; mostly clean, controlled audio; and a single language, single style bias under which detectors can appear strong yet fail to generalize.

*ASVspoof 2017.* This edition focused exclusively on the physical access (PA) scenario, addressing replay attacks recorded under controlled acoustic conditions (Kinnunen *et al.*, 2017). The corpus, derived from the RedDots data set, contained bona fide and replayed speech captured with a range of microphones, loudspeakers and environments, making it one of the first standardized resources for replay detection. Strengths: establish a benchmark for PA with systematically varied replay configurations that underscore the practical threat of replay attacks to speaker verification. Limitations: the diversity of devices and acoustic conditions was still limited compared to real-world replay scenarios, and the absence of other spoofing modalities (e.g. TTS, VC or deepfakes) restricted its generalizability beyond replay detection.

*ReMASC 2019.* Designed to approximate physical replay in everyday settings, ReMASC provided approximately 54,700 audio clips from 50 speakers with variations in devices, rooms and playback-capture configurations. Strengths: the emphasis on real acoustic channels, reverberation and device effects made it valuable for physical access studies and stress testing anti-replay front ends. Limitations: modest size and speaker count, limited linguistic diversity, and a focus on replay alone (not synthetic TTS/VC), which restricts its utility for modern neural deepfakes.

*ASVspoof 2019.* A landmark edition that scaled to more than 360,000 utterances and split evaluation into logical access (LA) (neural and statistical TTS/VC) and physical access (PA) (replay), while introducing tDCF to couple countermeasures with an ASV system (Todisco *et al.*, 2019). Strengths: task decomposition (LA vs PA) clarified problem settings; size and attack diversity improved statistical power; tDCF aligned research with deployed pipelines. Limitations: primarily monolingual and read speech; generator pipelines were fixed and documented, encouraging fingerprint chasing; and despite scale, later neural advances (diffusion, codec LMs) were not represented.

*FoR 2019.* The FoR data set contains 198,000 clips in English, designed to support spoofing detection research (Reimao and Tzerpos, 2019). Strengths: its size makes it practical for ablation studies and rapid prototyping, and it provides a substantial number of both bona fide and synthetic utterances. Limitations: attack type coverage is limited, the recordings are relatively short and clean, and the controlled conditions can inflate in-domain performance compared to real-world scenarios.

*WaveFake 2021.* Built to stress modern neural generation artifacts at scale, WaveFake emphasized coverage of contemporary TTS/VC models and vocoders with approximately 105,000–118,000 synthetic clips depending on the version. Strengths: focused pressure on neural artifacts, larger clip volume for data-hungry deep detectors, and straightforward train, validation and test splits. Limitations: limited channel diversity and recording conditions; dependence on a finite set of generators (risking overspecialization); and narrower linguistic variety relative to later multilingual corpora.

*FakeAVCeleb 2021.* A multimodal corpus containing audiovisual clips from 500 celebrities that couples manipulated video with synthetic or manipulated audio (Khalid *et al.*, 2021). The data set was developed with diverse ethnic backgrounds to address racial bias in deepfake data sets. Strengths: enables cross-modal and temporal consistency checks (lip-audio sync, identity coherence), reflecting real social media threat models; encourages fusion methods. Limitations: celebrity domain bias and potential copyright constraints; heterogeneous editing quality; and the risk that face and voice artifacts are data set-specific rather than general.

*ASVspoof 2021.* This edition broadened the challenge scope by introducing three tracks under a unified framework: LA, PA and speech deepfake (DF) detection. It also extended evaluation in English, providing over one million bona fide and spoofed utterances in total (Yamagishi *et al.*, 2021). Strengths: broader coverage of attack scenarios, and harder, more diverse spoofing lists that reduced the effectiveness of simple system fingerprinting. Limitations: despite the setup, many conditions (e.g. microphones, codecs and acoustic environments) remained relatively controlled, and emerging neural vocoder systems (e.g. diffusion-based models and codec language models) were not fully represented. In addition, the relatively short-duration utterances posed challenges for long-form and conversational detection research.

*ADD 2022.* A large-scale effort with more than 500,000 audio clips, designed to benchmark detection at an industrial scale. Strengths: sheer scale supports training data-hungry deep models, more granular demographic condition splits, and robust validation; broader speaker coverage improves identity variance. Limitations: licensing and redistribution constraints can limit academic reuse; language and content may be concentrated, leaving gaps for low-resource settings; and the compute footprint to exploit the full corpus is substantial.

*LibriSeVoc 2023.* A targeted corpus that “self-vocodes” clean speech with six popular neural vocoders (WaveNet, WaveRNN, MelGAN, Parallel WaveGAN, WaveGrad, DiffWave), containing approximately 92,400 audio samples (13,201 real and 79,206 synthetic) with 126.41 h of real audio, to foreground fingerprints of the decoder (Sun *et al.*, 2023). Strengths: controlled generation lets researchers isolate vocoder-level artifacts; supports multitask training for real and fake plus vocoder identification; and has shown resilience to standard postprocessing (resampling, additive noise). Limitations: focus on vocoder traces can underrepresent earlier acoustic modeling artifacts; reliance on read speech (LibriSpeech lineage) narrows stylistic coverage; and it does not include end-to-end diffusion or codec LM pipelines that may leave different cues.

*ASVspoof 5 2025.* The newest installment raised the bar with crowd-sourced deepfakes and adversarially crafted attacks that reflect realistic threat models. It broadened scoring beyond EER and tDCF to minDCF and $C_{llr}$ for better operating point and calibration assessment (Wang *et al.*, 2025). Strengths: closer alignment with malicious real-world generation, richer metrics for deployment-oriented evaluation and stronger cross-domain tests. Limitations: crowd-sourcing introduces label quality and provenance challenges; evolving, partially undisclosed attack pipelines can complicate reproducibility; and evaluation complexity increases, making comparisons harder to interpret without careful analysis.

Collectively, these data sets trace the evolution of the field: from early statistical spoofing (ASVspoof 2015), to replay realism and environmental variation (ReMASC, ASVspoof 2019), to neural deepfakes and multimodal settings (FoR, WaveFake, FakeAVCeleb, ASVspoof 2021), and most recently to large-scale, vocoder-focused and adversarially grounded resources (ADD, LibriSeVoc, ASVspoof 5). Table 7 provides a detailed breakdown of each corpus, summarizing its design focus, primary evaluation metrics, key advantages, limitations and release year. Building on this corpus-level analysis, Table 8 offers a direct comparison of closed-set (in-domain) versus open-set (out-of-domain) evaluation across these data sets. The advantages of this progression are clear: increasing scale, broader diversity of conditions and closer alignment with real-world deployment scenarios. However, the shortcomings are equally instructive: limited coverage of emerging diffusion- and codec-based generators, underrepresentation of spontaneous and noisy conversational audio, sparse cross-modal and long-form test cases, and persistent fragility under domain shift. We evaluate detector performance across three model groups: RawNet2 and LFCC-LCNN, both trained on ASVspoof; WavLM and Wav2Vec2-XLS-R, trained on alternative data sets; and RawNet2 vocoder, trained on LibriSeVoc. Future benchmarks should address these gaps by integrating modern generation methods, richer acoustic and linguistic variability, and standardized cross-corpus evaluation protocols. They should also establish more precise calibration targets to ensure that detectors trained under controlled conditions remain robust, scalable and trustworthy when deployed in the wild.

**Table 7.** Data sets and benchmarks for audio deepfake detection

| Name | Dateset size | Languages | Attack types | Evaluation metrics | Year |
|---|---|---|---|---|---|
| ASVspoof 2015 (Wu *et al.*, 2015) | 260k+ | English | VC, statistical TTS | EER | 2015 |
| ASVspoof 2017 (Kinnunen *et al.*, 2017) | 18k+ | English | Replay attacks | EER | 2017 |
| ReMASC (Reynolds *et al.*, 2019) | 55k+ | English | Replay, manipulation | EER, ROC-AUC | 2019 |
| ASVspoof 2019 (Todisco *et al.*, 2019) | 360k+ | English | Logical access (VC/TTS), physical access (replay) | EER, t-DCF | 2019 |
| FoR (Reimao and Tzerpos, 2019) | 198k | English | TTS, VC, replay | Accuracy, EER | 2019 |
| WaveFake (Müller *et al.*, 2021) | 105k-118k | English | Neural TTS, VC (multiple architectures) | EER, accuracy | 2021 |
| FakeAVCeleb (Khalid *et al.*, 2021) | 500 celebrities | English | Audio-visual deepfakes (TTS + face manipulation) | Accuracy, EER | 2021 |
| ASVspoof 2021 (Yamagishi *et al.*, 2021) | 500k+ | English | Logical access, physical access, deepfake | EER, t-DCF | 2021 |
| ADD (Yi *et al.*, 2022) | 500k+ | English, Mandarin | TTS, VC, hybrid | EER, accuracy | 2022 |
| LibriSeVoc (Sun *et al.*, 2023) | 90k+ | English | Vocoder artifacts | EER, robustness | 2023 |
| ASVspoof 5 (Wang *et al.*, 2025) | 1M+ | English | Crowdsourced deepfakes, adversarial attacks | EER, t-DCF, mindcf, $C_{llr}$ | 2025 |

**Table 8.** A compact meta-table comparing closed-set versus open-set, the results in EER (%), and are supported by Sun *et al.* (2023)

| Methods | Trained on | LibriSeVoc | WaveFake | ASVspoof |
|---|---|---|---|---|
| LFCC-LCNN (Lavrentyeva *et al.*, 2019) | ASVspoof | 0.14 | 0.19 | 11.60 |
| RawNet2 (Tak *et al.*, 2021b) | ASVspoof | 0.17 | 0.32 | 6.10 |
| WavLM (Chen *et al.*, 2022b) | Others | 0.45 | 2.92 | 6.94 |
| Wav2Vec2-XLS-R (Arun Babu *et al.*, 2021) | Others | 1.54 | 2.33 | 13.48 |
| Rawnet2 vocoder (Sun *et al.*, 2023) | LibriSeVoc | 0.13 | 0.19 | 4.54 |

## 5. Conclusion and future directions

AI-generated human voices have advanced rapidly in recent years, and their increasing realism has introduced significant challenges when exploited for malicious uses such as fraud and disinformation campaigns. This survey has provided a systematic overview of current AI voice synthesis techniques and the corresponding detection approaches. Yet, the fast pace of progress in generative models means that the arms race between generation and countermeasures is far from over. It has increasingly taken on a cat-and-mouse character, where improvements in synthetic voice generation prompt new detection methods, and each advance in detection then pushes generative models to become more challenging to catch. Looking ahead, we identify several broad trends and directions that are likely to shape both the generation and detection of synthetic audio, as well as the development of broader protective strategies, in the years to come.

We anticipate that next-generation speech synthesis will focus on enhanced data efficiency, greater linguistic flexibility and increased natural expressiveness. One prominent direction is *zero-shot or few-shot VC*, where models can capture a new speaker's voice from only a few seconds of example audio (Wang *et al.*, 2023). Recent neural language models, such as VALL-E, demonstrate that high-quality personalized speech can be generated using just a short acoustic prompt. Another frontier is *cross-lingual voice generation*, which enables the production of fluent foreign speech using a speaker's voice without parallel bilingual data (Jia, 2019). This leverages language-agnostic representations to preserve speaker identity across languages. In addition, *expressive and emotional speech synthesis* is gaining attention, using latent prosody spaces or style conditioning to mimic human-like intonation and affect (Skerry-Ryan *et al.*, 2018). Meanwhile, *diffusion-based voice models* offer new promise by iteratively refining waveforms from noise, resulting in greater stability and prosodic richness (Kong *et al.*, 2021).

On the detection front, efforts are increasingly focused on robustness and generalization. *Signal-level detectors* must now withstand real-world degradations such as compression and intentionally injected environmental noise (Tak *et al.*, 2021b). Beyond this, *phonetic and prosodic analysis* holds promise by leveraging high-level speech characteristics like coarticulation and intonation, which remain difficult for synthetic systems to replicate (Zhang *et al.*, 2018). *Multimodal detection* approaches, combining audio with video or metadata, are also emerging as more comprehensive tools for verifying authenticity (Mittal *et al.*, 2020). Crucially, future detection models must *generalize to unseen synthesis methods*, possibly through meta-learning or anomaly detection, as traditional classifiers often overfit to known attack types (Wu *et al.*, 2015).

Looking beyond detection, there is a growing recognition that preventative tools are necessary. *Watermarking and provenance tracking* techniques are being developed to embed imperceptible, verifiable signatures into synthetic speech (Lee, 2023). Complementary to this, *real-time spoof prevention* models are being optimized for deployment in live communication channels, where latency and computational cost are critical constraints (Todisco *et al.*, 2019). In parallel, *privacy-preserving verification* methods are evolving to support secure authenticity checks without exposing sensitive speaker data (Pathak *et al.*, 2021), an important consideration in both detection and legitimate applications of synthetic voice.

## Author contribution

Chengzhe Sun, Tianle Yang and Siwei Lyu contributed equally to this work.

## Disclaimer

This work represents opinions of the authors but not the funding agencies.

**Corresponding author**
Siwei Lyu can be contacted at: siweilyu@buffalo.edu